\documentclass[runningheads]{llncs}
\usepackage[T1]{fontenc}
\usepackage{graphicx}
\usepackage[inkscapelatex=false]{svg}
\usepackage{array} 
\usepackage{multirow}
\usepackage{tabularx} 
\usepackage{float} 
\usepackage{makecell}
\usepackage{placeins} 

\usepackage{hyperref}
\usepackage{adjustbox}
\usepackage{array,graphicx}

\begin{document}
\title{Classification of Raw MEG/EEG Data with Detach-Rocket Ensemble: An Improved ROCKET Algorithm for Multivariate Time Series Analysis
}

\titlerunning{Raw M/EEG Data Classification with Detach-Rocket Ensemble}

\author{Adrià Solana\inst{1}\inst{2}\and
Erik Fransén\inst{1}\inst{3}\inst{4} \and
Gonzalo Uribarri\inst{1}\inst{3}\inst{4}}

\authorrunning{A. Solana et al.}

\institute{
Section of Computational Brain Science, EECS, KTH Royal Institute of Technology, Stockholm, Sweden\\
\email{\{adriasic,erikf,uribarri\}@kth.se}
\and
Polytechnic University of Catalonia, Barcelona, Spain
\email{adria.solana@estudiantat.upc.edu}
\and
DigitalFutures, KTH Royal Institute of Technology, Stockholm, Sweden
\and
Science for Life Laboratory, KTH Royal Institute of Technology, Stockholm, Sweden
}
\maketitle              
\begin{abstract}
Multivariate Time Series Classification (MTSC) is a ubiquitous problem in science and engineering, particularly in neuroscience, where most data acquisition modalities involve the simultaneous time-dependent recording of brain activity in multiple brain regions. In recent years, Random Convolutional Kernel models such as ROCKET and MiniRocket have emerged as highly effective time series classification algorithms, capable of achieving state-of-the-art accuracy results with low computational load. Despite their success, these types of models face two major challenges when employed in neuroscience: 1) they struggle to deal with high-dimensional data such as EEG and MEG, and 2) they are difficult to interpret. In this work, we present a novel ROCKET-based algorithm, named Detach-Rocket Ensemble, that is specifically designed to address these two problems in MTSC. Our algorithm leverages pruning to provide an integrated estimation of channel importance, and ensembles to achieve better accuracy and provide a label probability. Using a synthetic multivariate time series classification dataset in which we control the amount of information carried by each of the channels, we first show that our algorithm is able to correctly recover the channel importance for classification. Then, using two real-world datasets, a MEG dataset and an EEG dataset, we show that Detach-Rocket Ensemble is able to provide both interpretable channel relevance and competitive classification accuracy, even when applied directly to the raw brain data, without the need for feature engineering.

\keywords{Multivariate Time Series Classification  \and EEG \and MEG \and ROCKET Algorithm}
\end{abstract}
\section{Introduction}
Multivariate Time Series Classification (MTSC) is a topic of increasing interest as the amount of acquired data and the number of possible applications in science and engineering grow over time. MTSC presents a challenging problem because a Multivariate Time Series (MTS) can exhibit complex patterns that not only span over time within a single channel, but also over time between multiple channels.



The standard approach in the Time Series Classification (TSC) literature is to develop complex architectures aimed at achieving high accuracy on established benchmarks \cite{bagnall2017great,ruiz2021great,middlehurst2023bake}, with the two most notable being the UCR (University of California, Riverside) archive for Univariate Time Series (UTS) and the UEA (University of East Anglia) archive for Multivariate Time Series (MTS) \cite{dau2019ucr,bagnall2018uea}. Examples of state-of-the-art models include TS-CHIEF \cite{shifaz2020ts}, InceptionTime \cite{ismail2020inceptiontime}, HIVE-COTE v2.0 \cite{middlehurst2021hive} and Hydra \cite{dempster2023hydra}. While these models are effective, their high accuracy is typically achieved at the expense of scalability, interpretability and computational feasibility.

Motivated by the lack of scalable models in the state-of-the-art, the RandOm Convolutional KErnel Transform (ROCKET) algorithms emerged as a lightweight alternative to TSC. This novel class of algorithms offers simple architectures capable of achieving accuracies comparable to much more complex models. There are three main variants of the ROCKET algorithm, namely ROCKET, MiniRocket and MultiRocket \cite{dempster2020rocket,dempster2021minirocket,tan2022multirocket}. Despite some differences, all of them share the same core idea. ROCKET works by generating a large number of random convolutional kernels, which are subsequently aggregated into features and used to train a ridge classifier \cite{dempster2020rocket}. Rather than relying on a learning process that fits the kernel parameters to the data, ROCKET depends on its extensive number of random features, with the expectation that some of them will carry information relevant to the classification task. Moreover, recent study demonstrates that it is possible to prune features of the model that are not relevant for classification through a process called Sequential Feature Detachment (SFD) \cite{uribarri2023detach}. The authors showed that the resulting pruned model, called Detach-Rocket, achieves better accuracy while requiring less than $10\%$ of the original number of features.

The ROCKET algorithm has proven successful in the TSC paradigm, achieving good performance on the UCR and UEA datasets, as well as other real-world applications \cite{uribarri2023deep}. The linear nature of its classifier helps to prevent overfitting on datasets with a limited number of instances and enables the model to handle large datasets efficiently due to its lightweight training scheme. However, there are two areas where the ROCKET model shows limitations:
\begin{itemize}
\item \textbf{Scalability with number of channels.} To deal with MTS, the original ROCKET algorithm combines all channels for each convolution. This means that each kernel has $L \times C+1$ coefficients, where $L$ is the length of the kernel and $C$ is the number of channels in the input time series. This simple multivariate strategy faces two problems when $C$ is large. First, the probability that a randomly sampled kernel will encode meaningful multichannel information decays rapidly as the number of interacting channels increases. This means that it will be increasingly difficult for the model to handle higher order interactions between channels as the number of channels grows. Second, even when the kernel encodes meaningful information for a subset of the channels, the contribution of the remaining channels to this convolution will become an obstacle to classification. These drawbacks have been partially addressed in MiniRocket and MultiRocket by limiting the number of channels mixed by each convolutional kernel, thus limiting the maximum possible order of channel interactions. A downside of this strategy is that each kernel now includes only a random subset of channels, which for large $C$ introduces a lot of variability between different realizations of the model and requires a very large number of kernels to properly sample all possible channel combinations.
\item \textbf{Interpretability.} Despite their simple architecture, ROCKET models are difficult to interpret. Some of the limitations in this regard are: a) the generated feature space is very high dimensional, b) there is no built-in way to have a label probability, and c) in the case of MTSC, there is no easy way to compute the channel importance (i.e., the relevance of each channel to the classification task). 

\end{itemize}
These limitations are particularly notable when the models are applied to neural data. Noninvasive brain activity is typically measured using inherently multivariate modalities, such as electroencephalograms (EEG) \cite{britton2016electroencephalography}, magnetoencephalograms (MEG) \cite{hamalainen1993magnetoencephalography}, and functional magnetic resonance imaging (fMRI) scans \cite{glover2011overview}. In most cases, these multivariate time series have a large number of channels representing either different sensors or regions of interest (ROIs). In addition, in neuroscience, it is usually important to identify which brain regions are most relevant for discriminating between the different classes, defined typically as experimental conditions, groups of participants, etc. Furthermore, when the model is used for diagnosis, having a reliable measure of label probability is essential for evaluating the robustness of the classifier and for setting a threshold that balances the ratio of false positives to false negatives. Thus, even though ROCKET models have shown some potential for neuroscience tasks \cite{rushbrooke2023time,mizrahi2024comparative}, there are fundamental limitations of the methodology that need to be addressed for this type of data.

In this work, we introduce a novel ROCKET-based model, named Detach-Rocket Ensemble, specifically designed to address multivariate time series classification problems. Our methodology involves creating an ensemble of Detach-MiniRockets, which are pruned MiniRocket models. The ensemble mitigates model variability \cite{dietterich2000ensemble} and enables the exploration of a much larger pool of kernels than a single MiniRocket, increasing the probability of finding meaningful interactions between channels. Due to the small size of the pruned Detach-MiniRocket models, the resulting ensemble is typically smaller than a single MiniRocket. Another advantage of our model is that, as an ensemble, it provides a built-in measure of label probability. In addition, by leveraging the pruning process of Detach-Rocket, we have also developed a straightforward method for estimating channel relevance for the classification task. 

To evaluate the channel relevance estimation of our model, we first designed a synthetic MTSC dataset for which we are able to control the amount of information about the classification label carried by each channel of a multivariate time series. We show that our methodology is able to identify the relevant channels and their relative importance for the classification task.

We then demonstrate the potential of our model on two real-world neuroscience applications. The first consists of a face recognition task using 306-channel MEG data. We show that our model is able to achieve better accuracy on this MTSC dataset than previous ROCKET-based models, and is also able to identify the relevant brain areas for classification, which are consistent with those found in the neuroscience literature for the same task. The second is an Alzheimer's disease classification task using resting-state EEG. We show that our model significantly outperforms state-of-the-art models designed for raw EEG classification, achieving an accuracy comparable to that obtained by applying feature engineering specifically designed for this task. For this classification task, we compute the ROC curve of our model and select the optimal threshold, demonstrating the benefits of incorporating a label probability.

The rest of the paper is organized as follows. In Section 2, Background, we review previous ROCKET algorithms relevant to this work, with a particular focus on Detach-Rocket. In Section 3, Methods, we introduce the Detach-Rocket Ensemble model and the methodology for estimating channel relevance. In Section 4, Synthetic Dataset, we present the results of the synthetic dataset experiment. In Section 5, Real Datasets, we demonstrate the application of the Detach-Rocket Ensemble to EEG and MEG datasets. Section 6 contains the discussion, and Section 7 presents the conclusions of this study.




\section{Background}
\subsection{Random Convolutional Kernel Transform (ROCKET)}
The standard ROCKET model use 10,000 random convolutional kernels to compensate for the lack of a training process typically used on architectures such as Convolutional Neural Networks (CNNs) \cite{alzubaidi2021review}. These kernels convolute the normalized input time series to extract informative features. To achieve this, the kernel parameters such as the length, weights and dilation are drawn from random distributions.

The convolution of each kernel over a time series results in a feature map, equivalent to a univariate time series that encodes the extracted characteristics. Each of these feature maps are pooled into two scalars through two operators: global max pooling, that selects the maximum value, and the Percentage of Positive Values (PPV), that computes the fraction of values larger than zero. The 20,000 resulting features fit a classifier, usually based on a ridge regression model, that yields a label prediction for the given time series.


\subsection{MiniRocket}
MiniRocket \cite{dempster2021minirocket} was developed as a faster and lighter version of the original ROCKET algorithm. This enhanced architecture is more deterministic when generating the kernels and also eliminates the max pooling operator, thus relying solely on the 10,000 PPVs. These modifications simplify the model and yield slightly superior performance compared to ROCKET in the UCR archive \cite{dempster2021minirocket}.

The MiniRocket parameter selection is fully deterministic, except for the bias calculation and the channel selection. The set of biases to be used in prediction is computed after an initial convolution of a subset of input samples, where each bias is obtained as a quantile from the resulting feature maps. Hence, fully deterministic biases can be obtained if the full training set is used.

For MTS, MiniRocket chooses for each kernel combinations from 1 up to 9 channels with an exponentially decaying distribution, i.e. convolutions are more likely to use less number of channels. The channels that are used for each kernel are chosen with equal probability. In the context of our work, this channel sampling is crucial for studying the relation between the best performing features and the channels that generated them, which is key for the interpretability of MTS.

\subsection{Arsenal}
The state of the art of TSC is currently dominated by HIVE-COTE v2 \cite{middlehurst2021hive}, a meta ensemble that aggregates different feature extraction and classification modules into a single prediction, achieving cutting-edge accuracies \cite{middlehurst2023bake}. One of these modules is Arsenal, an ensemble of 25 ROCKET models with 2,000 kernels each. Although Arsenal serves as a component within the full model, it can function as a standalone classifier. In this case, the algorithm works by cross-validating every ROCKET model with the training dataset and producing estimates via soft-voting the predictions, weighted by the scores of each model during cross-validation. The ensemble methodology aims to improve the predictive power of the classifier as well as giving a probability estimate. 

\subsection{Detach-Rocket}
\label{sec:detach}
\begin{figure}
\includegraphics[width=0.9\textwidth]{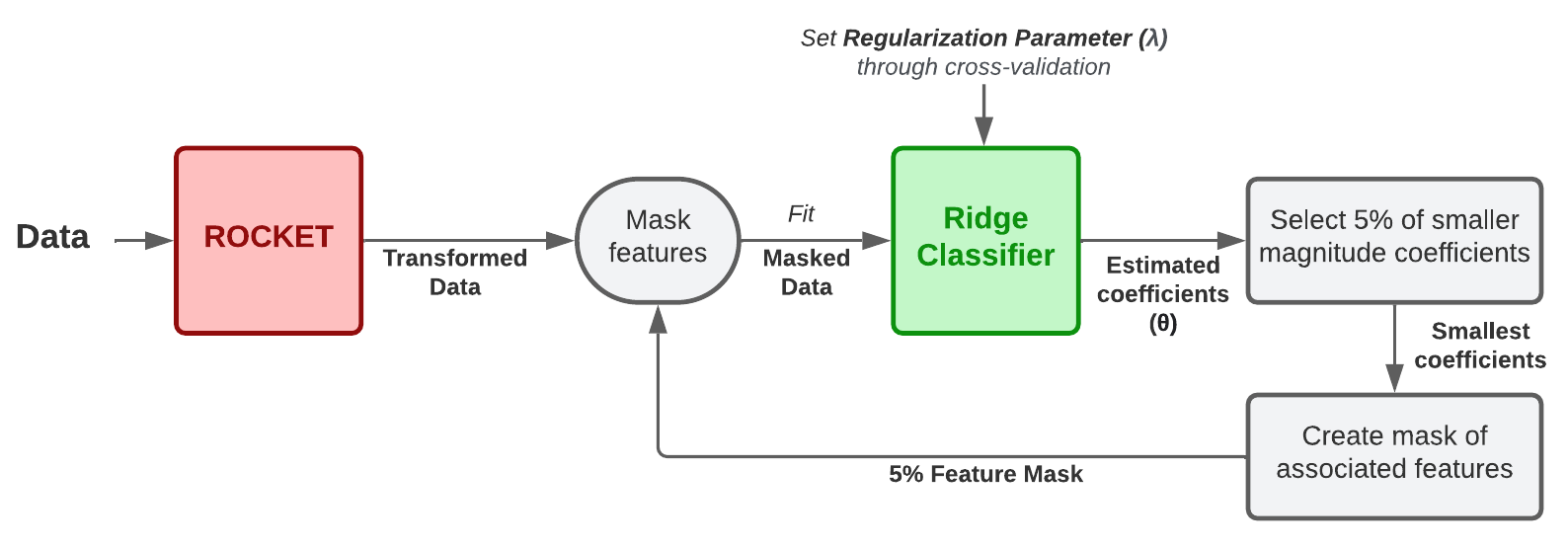}
\centering
\caption{Sequential Feature Detachment (SFD) diagram.} \label{fig:SFD}
\end{figure}

ROCKET provides a large number of features with the objective of training a simple linear classifier for a relatively low cost. However, there is no straightforward method to decide which features are truly relevant, and which ones solely add undesired complexity or even hinder the classification. A previous study presented Sequential Feature Detachment (SFD) \cite{uribarri2023detach}, a novel algorithm that intends to prune features by keeping the most essential ones (Figure \ref{fig:SFD}). It does so by iteratively fitting a ridge classifier, pruning a percentage of the features associated with the coefficients of lesser magnitudes, and fitting it again until the desired percentage of features is achieved.

In this study, the efficacy of this methodology was evaluated on all binary classification datasets from the UCR archive. The results demonstrated that a full ROCKET model could be pruned to 2\% of its original features while maintaining overall classification accuracy \cite{uribarri2023detach}. These findings indicate that SFD can effectively reduce the complexity of a model while enhancing its generalization capabilities.

Furthermore, the authors proposed an end-to-end model named Detach-Rocket that automatically determines how many features should be kept in a pruned model. This method employs a trade-off hyperparameter $c$, which weights the retention of features against the accuracy of the pruned model. As $c$ approaches 0, greater significance is given to accuracy, while larger values prioritize maintaining a smaller size. The default value suggested for $c$ is $0.1$, which gives more importance to accuracy and prunes a model down to roughly 1\% in mean according to the tests on the UCR archive.

\vspace{-1.5em}
\section{Methods}

\subsection{Detach-Rocket Ensemble Model}

\subsubsection{Motivation.}
As discussed in the Introduction section, when the number of channels $C$ of an MTS is high, ROCKET-based models require a larger sample of convolutional kernels to properly sample the space of possible patterns. One way to do this is to simply increase the number of kernels in a single model. While this strategy may be effective, it is computationally inefficient. First, the forward pass or inference in the model will become slower as the number of convolutions required increases. Second, for a very large number of kernels, it will be impossible to fit the entire feature matrix of size (num. instances, num. features) into RAM, making training either much slower or infeasible. 

As a solution to this problem, we introduce Detach-Rocket Ensemble. Instead of training a single model with a large number of kernels, we propose to train several standard models and then aggregate them into an ensemble. This makes the training process both manageable in terms of feature matrix size and trivially parallelizable, while still enabling the exploration of a large pool of kernels.

A key aspect of Detach-Rocket Ensemble is that it is composed of pruned Detach models. These pruned models have been shown to achieve the same classification accuracy as the full model while retaining only $2\%$ of the original features \cite{uribarri2023detach}. Consequently, although a large number of kernels are explored during training and pruning, the total number of kernels in the resulting ensemble model is comparable to that in a single standard ROCKET, thereby requiring approximately the same convolutions for inference.

In addition to potentially improving the classification accuracy, an ensemble model provides a reliable class probability, which is useful to explore different thresholds that may better suit the classification task. Furthermore, in our case, the ensemble also helps to achieve a better estimation of channel relevance, as discussed in the next section.

\subsubsection{Model description.}
\begin{figure}
\includegraphics[width=0.8\textwidth]{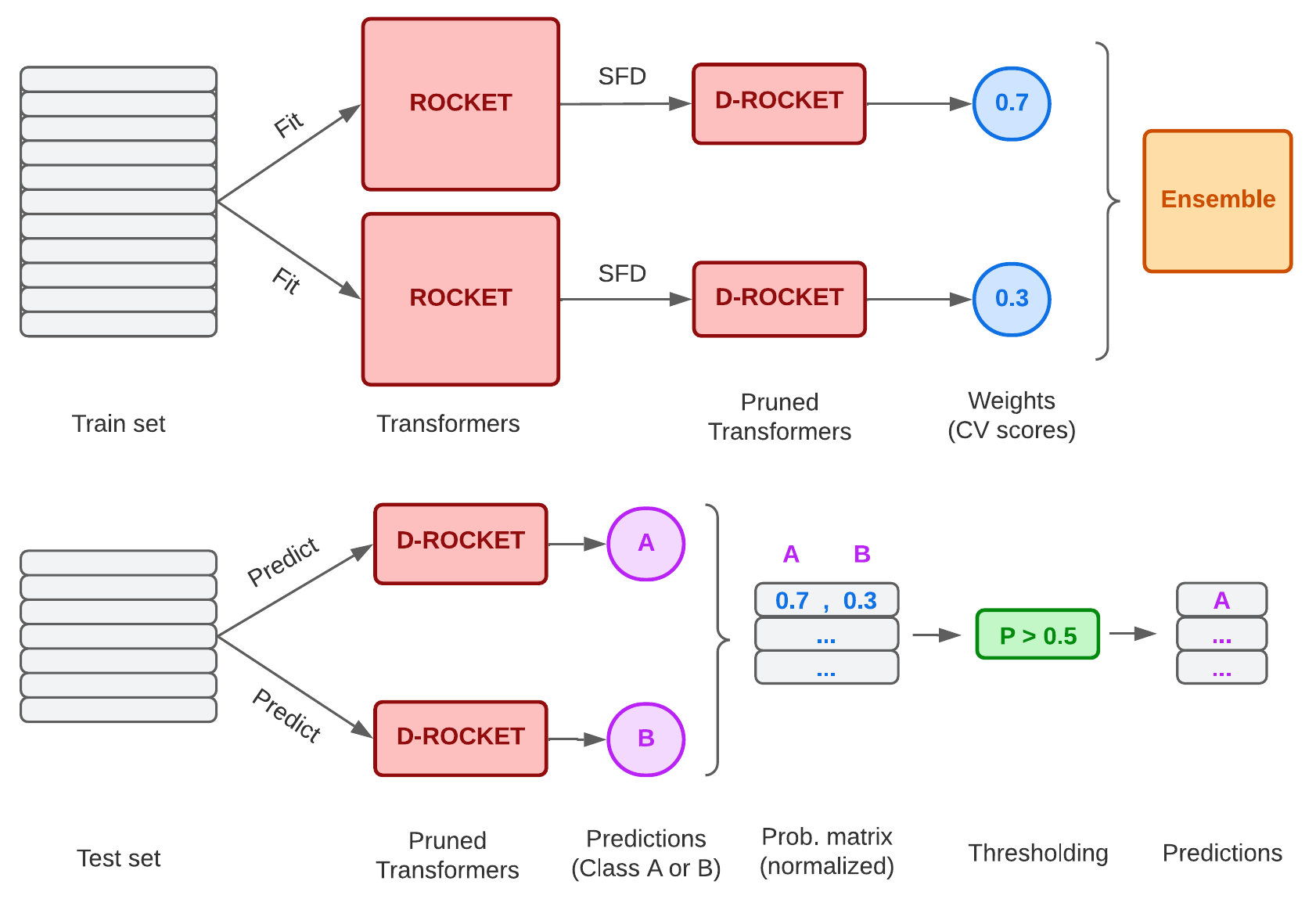}
\centering
\caption{Methodology of the Detach-Rocket Ensemble, fit (top) and predict (bottom) with N = 2 models and two classes (A and B). D-Rocket is the abbreviation of a single Detach-MiniRocket classifier.} \label{fig:methodology_ensemble}
\end{figure}

A Detach-Rocket Ensemble model is composed of $N$ Detach-MiniRocket models, which are first trained independently. The training process for each model follows the procedure described in Section \ref{sec:detach}: Each MiniRocket is pruned with SFD, using a subset of the training set to determine the optimal pruning size. Each of the resulting Detach-MiniRockets is then assigned a weight in the ensemble according to its performance on the training set. A schematic of this training process for a Detach-Rocket Ensemble model with $N=2$ number of models is presented in Figure \ref{fig:methodology_ensemble}.

To perform a prediction on a given instance, the model weights the label prediction made by each Detach-MiniRocket with the corresponding model weight and generates a normalized label probability. This probability can be subsequently thresholded to obtain the final model decision on that instance. The prediction process for a Detach-Rocket Ensemble model with $N=2$ is also illustrated in Figure \ref{fig:methodology_ensemble}.

In the present work, we chose Detach-MiniRocket as the base classifier for our ensemble because it is the fastest ROCKET variant. However, the exact same methodology can be applied using Detach-MultiRocket, which, given its stronger solo performance, may yield even better results for some datasets. It is important to note that, for large $C$, the variability across the learners arises mainly from the random subset of channels selected for each of the kernels. Although it is possible, we do not recommend using the standard ROCKET as the base classifier due to the reasons discussed in the introduction section and its slower training time.

\subsection{Channel Relevance Estimation}

\begin{figure}
\includegraphics[width=0.8\textwidth]{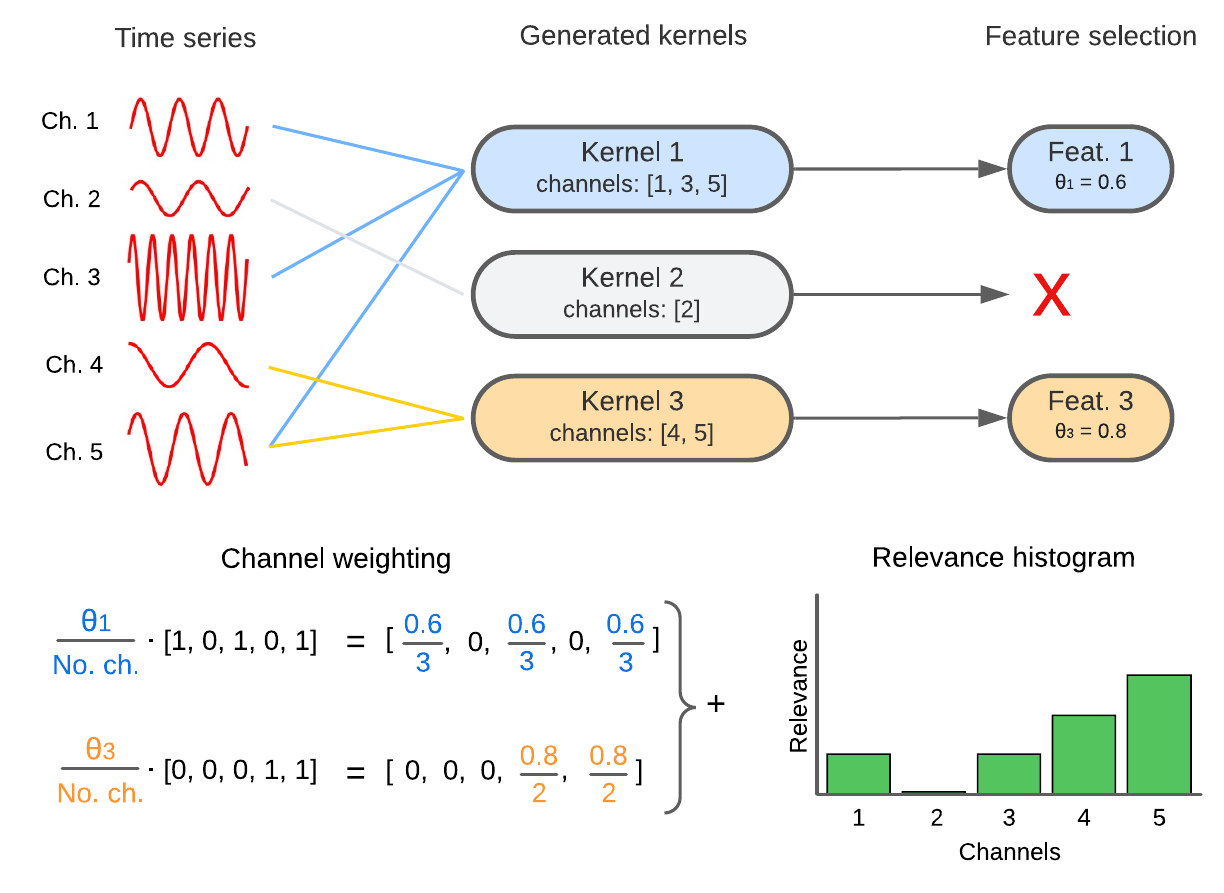}
\centering
\caption{Methodology employed to estimate the relevance of each channel using Detach-Rocket. In this example, three kernels are applied over a time series comprising five channels. Subsequently, SFD selects the first and third features, thereby selecting the first and third kernels. The channels combined by these kernels are then weighted and accounted for in the computation of the channel relevance histogram.} \label{fig:method}
\end{figure}

To estimate the feature relevance in a Detach-Rocket Ensemble model, we first compute it on each of its constitutive base classifiers. The methodology for estimating relative channel relevance on a single Detach-MiniRocket is illustrated in Figure \ref{fig:method} and consists of the following steps:
\begin{enumerate}
\item SFD selects the most relevant kernels in the MiniRocket model.
\item The model retrieves the channels that each selected kernel has used.
\item Each channel is given an importance proportional to the weight ($\theta_i$) of its kernel divided by the number of channels in the same kernel.
\item The pondered channels are summed and normalized into a relative relevance histogram.
\end{enumerate}

The pruning process is the essential part of this methodology, as it selects the relevant kernels and completely removes the irrelevant ones from the channel importance calculation, which is the majority of them. The subsequent weighting, based on the coefficient of the ridge classifier and the number of channels involved, is a way to achieve a greater sensitivity in the identification of the relevant channels. This methodology has been developed and optimized through a series of tests on a fully controlled synthetic dataset. 

Once the relative channel relevance has been computed for all base classifiers, the resulting channel relevance for the ensemble model is determined by taking the median of the relevancies obtained for each channel in the base models and normalizing along the channels. In addition to potentially improving classification accuracy, the ensemble mitigates the variability in the base models' channel relevance estimates, resulting in a more reliable estimate.

Note that the proposed channel relevance estimation is built-in, meaning that it does not require additional computation for strategies such as predicting instances with masked channels or retraining the algorithm on a subset of channels.





\section{Synthetic Dataset}
\subsection{Synthetic Dataset Design}
In this first experiment, we created a synthetic time series classification dataset to test our proposed channel relevance estimation procedure. The goal was to have a controlled environment where the contribution of each channel to the classification task is known, so that we can compare the estimated channel relevance to a ground truth.

\begin{figure}[ht!]
\includegraphics[width=0.9\textwidth]{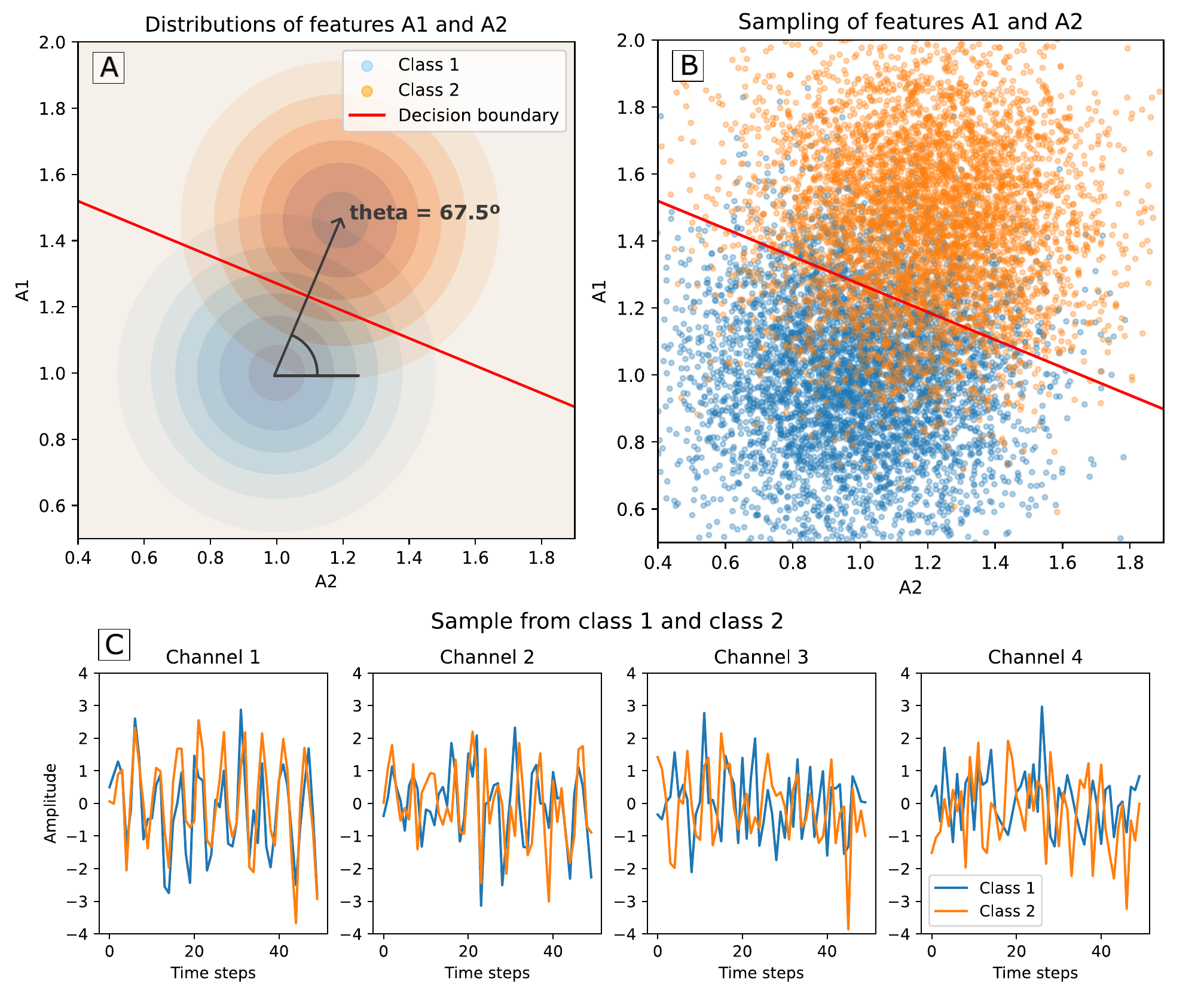}
\centering
\caption{Example plots depicting the synthetic dataset. Plot A shows the Gaussian distributions from which the sinusoidal amplitudes of channels 1 and 2 (A1 and A2) are sampled, for both classes, in addition to the analytical ideal decision boundary. It also shows the effect of theta, that shifts the center of the distribution of class 2 around class 1. Plot B shows a sampling instance of these distributions. Plot C presents one sample of each class, both unraveled along their 4 channels, after the amplitudes have been sampled and white noise has been added.} \label{fig:synthetic_dataset_method}
\end{figure}

To this end, we constructed a four-dimensional binary time series dataset. This synthetic dataset is designed so that one parameter, theta ($\theta$), controls how the information for classification is distributed along two of the four channels. The remaining two channels are noninformative channels used as control. Specifically, the description of the channel content is as follows: Channels 1 and 2 contain two different sinusoidal waveforms whose amplitude values are relevant to the classification task, Channel 3 contains a sinusoidal waveform whose amplitude is not relevant to the classification task, and Channel 4 contains no waveform. All channels also contain additive white noise. Figure \ref{fig:synthetic_dataset_method} (C) presents a representation of two samples of this dataset, one from each of the classes.

For a given value of theta, the dataset is created by sampling the values of the sinusoidal amplitudes A1 and A2 in channels 1 and 2 respectively from a pair of Gaussian distributions, one belonging to each of the two classes. Theta controls the positioning of these distributions in the parameter plane, the plane of all possible sinusoidal amplitudes as depicted in Figure \ref{fig:synthetic_dataset_method} (A). See also Figure \ref{fig:synthetic_dataset_method} (B) for a particular sampling instance. The theta angle controls the relative position of the Gaussian distributions with respect to the amplitude plane, thus determining the amount of information about the classification task carried by each of the amplitudes. If theta is $45^\circ$, then the class label information is equally distributed between channels 1 and 2. If theta is zero, then all the information about the class is carried by channel 2, while if theta is $90^\circ$, all the information is carried by channel 1. 

To estimate the difficulty of this classification task, we calculated the optimal decision boundary, depicted in red in Figure \ref{fig:synthetic_dataset_method} A and B. This determines the maximum achievable accuracy when classifying in the parameter space, constituted by the first and second channels amplitudes. The analytical results demonstrated that, for all values of theta, this accuracy was $84.13\%$. The complexity of the actual classification task is much higher, since these amplitude parameters are then transformed into waveforms, two interfering channels are included, and high-power white noise is added to the time series. Figure \ref{fig:synthetic_dataset_method} C shows two instances of the synthetic dataset that illustrate the non-triviality of the classification task.



\subsection{Synthetic Dataset Results}
We explored the relevance of the channels for several values of theta ranging from $0^\circ$ to $90^\circ$. For each one of the resulting datasets, we used a Detach-Rocket Ensemble of 25 models with 10,000 kernels each to estimate the channel relevance. The dataset design is such that the first channel -with a sinusoid of amplitude A1- is irrelevant when $\theta=0^\circ$, and that it increases with theta until it is the only important channel at $\theta=90^\circ $. The opposite happens with the second channel, with a sinusoid of amplitude A2. While the relevance values of the channels are not strictly defined at intermediate thetas, they should be equal at $\theta=45^\circ$, and they should monotonically increase/decrease.

\begin{figure}[ht!]
\includegraphics[width=0.9\textwidth]{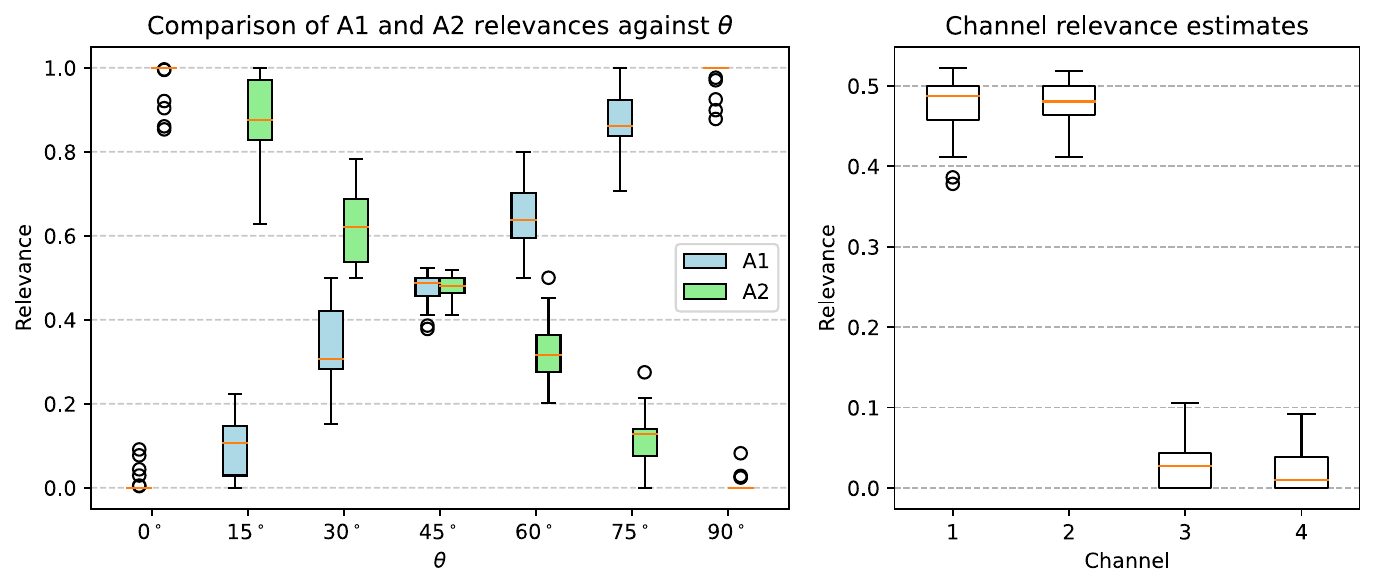}
\centering
\caption{Box plots of the synthetic dataset channel relevance estimates obtained with 25 Detach-MiniRocket models. The plot in the left shows the estimates of channels 1 and 2 (with sinusoidal amplitudes A1 and A2) along 7 different values of theta. For a given value of theta, box plots are presented side by side for better visibility. The right plot shows the estimates for all four channels when theta = $45^\circ$. Both figures are normalized so that the total relevance value adds up to one, and the orange lines indicate the median.} \label{fig:synthetic_dataset_results}
\end{figure}

In Figure \ref{fig:synthetic_dataset_results}, the left plot presents the estimated importance for channels 1 and 2 as a function of the angle theta. The estimation of relevance of the channels is expressive and captures how the importance shifts from channel 2 to channel 1 when the angle theta changes from $0^\circ$ to $90^\circ$. Note that the relevance estimates made by the individual Detach-MiniRocket models present exhibit some variance, but the median of the distributions correctly captures the expected behavior. This shows how the ensemble nature of the model also helps to get a better estimate of the channel importance. In Figure \ref{fig:synthetic_dataset_results}, the right plot shows the estimated relevance for all four channels when theta equals $45^\circ$. It can be observed that, even though the model assigns some relevance to the unimportant third and fourth channels, it clearly identifies that the first and second ones are the most important in an equal proportion.

\vspace{-1em}


This experiment demonstrates that the Detach-Rocket Ensemble provides an accurate estimate of the channels that are more relevant for classification in this controlled environment. However, despite the effectiveness of our synthetic dataset in validating the methodology, it may not fully reflect the performance of the algorithm in complex real-world datasets, where factors such as high-dimensionality and low class separability may limit its ability to properly estimate channel relevance. In the following section, we illustrate the potential of our procedure in real-world neuroscience applications.

\section{Real-world Datasets}
\subsection{Face Detection (MEG)}
\subsubsection{Dataset Description.}
We first tested our methodology on a MEG dataset collected and defined in \cite{henson2011parametric} and shared in a Kaggle competition \cite{decoding-the-human-brain}. The data were acquired with an array containing 306 sensors: one magnetometer and two orthogonal planar gradiometers at each of the 102 positions across the scalp. The signals were sampled at 1.1kHz and low-pass filtered with a cut-off frequency of 350Hz. Using this configuration, data were collected from subjects who were presented with either pictures of faces or pictures of scrambled faces for a duration of less than one second. The proposed task consisted of classifying trials in which the subject observed a regular face against trials in which the subject was presented with a scrambled face.

\subsubsection{Results.}
This dataset was used to test the performance of Detach-Rocket Ensemble in terms of both accuracy and channel relevance estimation. To evaluate its accuracy, we compared it to three other ROCKET-based models: MiniRocket, Detach-MiniRocket (abbreviated as D-MiniRocket), and Arsenal. To evaluate the channel relevance estimation, we compared it with previous analysis of relevant brain regions for the same task.

First, a hyperparameter optimization was conducted to determine the number of kernels that produced the most accurate results for MiniRocket, Detach-Rocket, and the Detach-Rocket Ensemble. To do this, 3 out of the 16 subjects present in the train set were used as a validation set on which the scores were computed, and the remaining 13 were used for training. The Appendix table \ref{tab:kernel_optimization} shows the results for this exploration on all the models. The validation accuracy showed an overall increase with the number of kernels, capped at 10,000 kernels in the case of the Detach models, but beneficial up to the maximum number of explored kernels for MiniRocket (20,000). The best performing strategy in this initial search was a Detach-Rocket Ensemble with 10,000 kernels. 


Given a specific dataset, further optimization of the Detach-Rocket Ensemble may improve the models performance. Some relevant parameters include the trade-off coefficient $c$, which tunes the extent to which the model is pruned, and the number of estimators $N$, which adds complexity to the ensemble. For this particular experiment we use $c=0.1$, which is the default value for Detach-Rocket, and $N=25$, to match the number of estimators of Arsenal's architecture.




We also evaluated the performance of Arsenal on the classification task, but we had to implement an alternative version using MiniRocket as base model instead of the default ROCKET. The reason for this was that the large number of channels present in the MEG dataset makes the default implementation unfeasible to run in our computational environment.

\begin{figure}[ht]
    \centering
    \begin{minipage}{0.45\textwidth}
    
        \centering
        \renewcommand{\arraystretch}{2}
        \setlength{\tabcolsep}{5pt}
        \begin{adjustbox}{width=\columnwidth,center}
        \begin{tabular}{|c|c|c|}
            \hline
            \textbf{Model} & \textbf{Train (\%)} & \textbf{Test (\%)} \\
            \hline
            \makecell{MiniRocket\\(20k kernles)}& 80.2$\pm$0.2 & 59.7$\pm$1.5 \\
            \hline
            D-MiniRocket & 72.2$\pm$2.9 & 60.8$\pm$0.5 \\
            \hline
            Arsenal & 87.4$\pm$0.1 & 61.5$\pm$0.4 \\
            \hline
            \makecell{\textbf{D-Rocket}\\\textbf{Ensemble}} & 78.6$\pm$0.3 & \textbf{64.3$\pm$0.5} \\
            \hline
        \end{tabular}
        \end{adjustbox}
    \end{minipage}
    \hfill
    \begin{minipage}{0.5\textwidth}
        \centering
        \includegraphics[width=\textwidth]{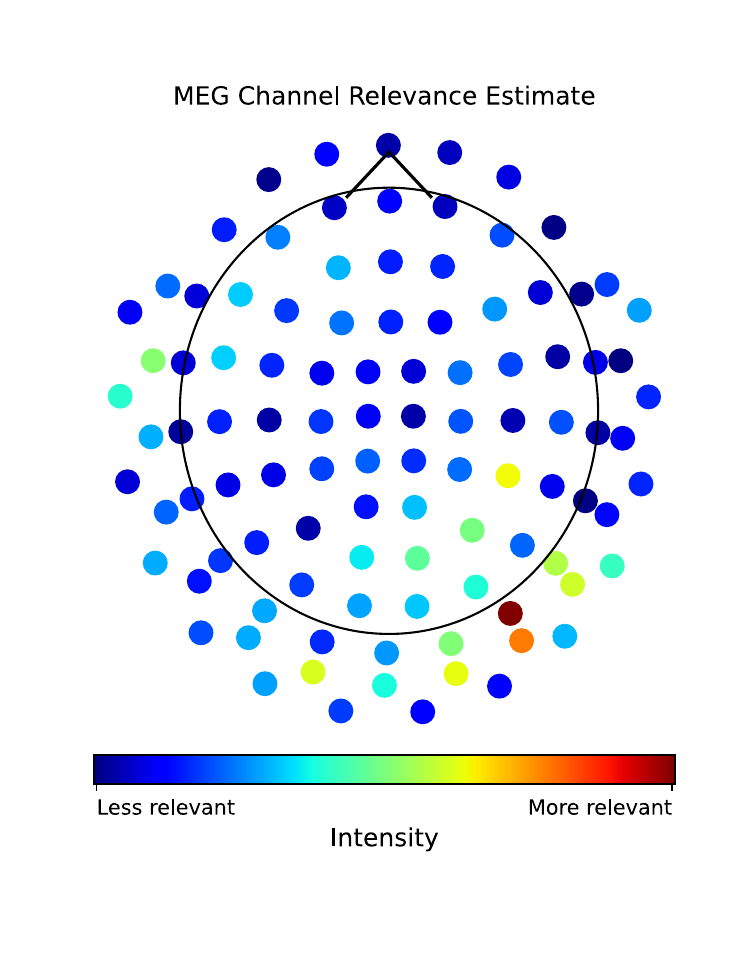}
        \vspace{-4em}
    \end{minipage}
    \caption{(Left) Train and test accuracies (mean $\pm$ standard deviation) obtained by running the optimal models three times. (Right) Mean channel relevance estimates of the Face Detection task  over three Detach-Rocket Ensembles.}
    \label{fig:FD_scores_and_relevance} 
\end{figure}

Finally, we compare the performance over three independent runs of the four models using the best configuration tested for each of them. The statistical results of these experiments are depicted in Figure \ref{fig:FD_scores_and_relevance} (left). Detach-Rocket Ensemble obtains the best test accuracies among the evaluated models, while also showing lower overfitting than Arsenal as a consequence of the feature pruning. On average, the feature detachment process pruned the models down to 6.2\% of their original features. In terms of average training time in our computational setting, a single MiniRocket model required 9.46 minutes, a Detach-MiniRocket model required 10.38 minutes, and the entire Detach ensemble required 263.66 minutes.

In Figure \ref{fig:FD_scores_and_relevance} (right), we present the results of the channel relevance estimation using Detach-Rocket Ensemble. To evaluate our methodology, we compared the results with those obtained in the study from which the data originated \cite{henson2011parametric}. In that study, the authors used statistical parametric mapping to find brain regions where MEG activity differed in recordings obtained from participants observing regular or scrambled faces. They found that the right lateral occipital cortex was the brain area that showed the most significant differences (see Figure 6, panel A, for MEG data in \cite{henson2011parametric}). This is precisely the same area that our methodology highlights as the most important for classification, demonstrating its ability to correctly estimate channel relevance in a complex high-dimensional real-world dataset.

\subsection{Alzheimer's Disease Classification (EEG)}
\subsubsection{Dataset Description.}
This section employs the dataset presented in \cite{data8060095}. The dataset comprises EEG recordings from 88 participants. A total of 36 individuals were diagnosed with Alzheimer's disease (AD), 23 were diagnosed with frontotemporal dementia (FTD), and 29 were classified as healthy control subjects (CN). The mean age of each group was 66.4 years, 63.6 years, and 67.9 years, respectively. For each participant, the recording was conducted in a resting state with the eyes closed. The recording device utilized 19 scalp electrodes (Fp1, Fp2, F7, F3, Fz, F4, F8, T3, C3, Cz, C4, T4, T5, P3, Pz, P4, T6, O1, and O2) with two reference electrodes (A1 and A2) located on the mastoids.  The preprocessing of these recordings included band-pass filtering within the range of 0.5 to 45 Hz and artifact correction using the Artifact Subspace Reconstruction routine (ASR) and Independent Component Analysis (ICA). The recording montage was referential using Cz for common mode rejection but, during the preprocessing, the signals were re-referenced to the average value of A1-A2.

\subsubsection{Results.}
In order to evaluate the effectiveness of our model, we compare its performance with that of the study presented in \cite{miltiadous2023dice}. The study, authored by the researchers who collected the dataset \cite{data8060095}, introduces the Dual-Input Convolution Encoder Network (DICE-net), a classifier tailored to classify AD in this particular data modality. DICE-net employs engineered biomarkers to train a convolutional and transformer-based architecture. We conducted our experiments on the AD vs. CN classification task, as this is the task on which the study focuses.


To match the validation methodology used in \cite{miltiadous2023dice}, we employed Leave One Subject Out (LOSO) cross-validation, conducting the experiment 65 times, one fold for each of the subjects. This process yielded a set of 65 small confusion matrices —each obtained by training the model on 64 subjects and predicting the trials of the remaining one— which were then summed to derive the final results. In diagnostic tasks of this nature, a careful validation scheme, where trials in the test set belong solely to subjects unseen during training, is crucial for accurately evaluating the model's generalization ability. Table \ref{tab:ad_cn_results} presents the results, using the same evaluation metrics as in \cite{miltiadous2023dice}. We split this table in two sections. The first shows models that require previous feature engineering, including DICE-net. The second part of this table shows the results for state-of-the-art deep learning models designed for raw EEG signal classification. Although not specifically designed for EEG, we include the Detach-Rocket Ensemble in this second part as it is also use raw EEG as input.




\begin{table}[ht]
\centering
\renewcommand{\arraystretch}{1.5}
\setlength{\tabcolsep}{10pt} 
\caption{AD vs CN scores reported by the models tested in \cite{miltiadous2023dice} and our of Detach-Rocket Ensemble (bottom, D-Rocket Ensemble). The table includes both models requiring feature engineering and models using the raw EEG signal. The presented metrics are Accuracy (ACC), Sensitivity (SENS), Specificity (SPEC), Precision (PREC) and F1-Score (F1).}
\setlength\extrarowheight{-1pt}
\begin{adjustbox}{width=0.8\columnwidth,center}
\begin{tabular}{|c|l|ccccc|}
\hline
\textbf{Type} &\textbf{AD/CN model} & \textbf{ACC} & \textbf{SENS} & \textbf{SPEC} & \textbf{PREC} & \textbf{F1} \\
\hline
\parbox[t]{2mm}{\multirow{7}{*}{\rotatebox[origin=c]{90}{Feature engineering}}}
&LightGBM    & 76.28\% & 76.08\% & 76.52\% & 79.67\% & 77.83\% \\
&XGBoost     & 75.53\% & 76.08\% & 74.87\% & 78.55\% & 77.29\% \\
&CatBoost    & 75.39\% & 75.50\% & 75.25\% & 76.68\% & 77.05\% \\
&SVM+PCA     & 73.75\% & 71.51\% & 76.46\% & 78.60\% & 74.89\% \\
&PCA-kNN     & 72.52\% & 70.30\% & 75.19\% & 77.41\% & 73.69\% \\
&MLP         & 73.69\% & 72.98\% & 74.81\% & 77.80\% & 75.31\% \\
&DICE-net \cite{miltiadous2023dice}    & 83.28\% & 79.81\% & 87.94\% & 88.94\% & 84.12\% \\
\hline

\parbox[t]{2mm}{\multirow{5}{*}{\rotatebox[origin=c]{90}{Raw EEG}}}
&EEGNet \cite{lawhern2018eegnet}                    & 41\%    & 47.20\% & 37.67\% & 37.89\% & 42.04\% \\
&EEGNetSSVEP \cite{waytowich2018compact}                & 51.46\% & 56.78\% & 45.39\% & 47.65\% & 51.82\% \\
&DeepConvNet \cite{schirrmeister2017deep}              & 54.21\% & 45.43\% & 57.59\% & 48.71\% & 47.01\% \\
&ShallowConvNet \cite{schirrmeister2017deep}             & 42.18\% & 46.50\% & 41.11\% & 49.74\% & 48.07\% \\
&\textbf{D-Rocket Ensemble} & 79.86\% & 78.89\% & 80.47\% & 74.89\% & 76.84\% \\
\hline
\end{tabular}
\end{adjustbox}
\label{tab:ad_cn_results}
\end{table}




Table \ref{tab:ad_cn_results} shows that our model significantly outperforms all alternatives using raw EEG data. These deep learning architectures designed for raw EEG signal classification overfit the training set and fail to generalize, achieving an accuracy no better than chance for unseen subjects. In fact, our model performs better than most of the models using feature engineering, being outperformed in accuracy only by DICE-net. Moreover, when implementing majority voting on subjects' trial predictions to obtain a class label per subject, our model achieves a subject-level accuracy of $86.15\%$, better than the $84.62\%$ obtained by DICE-net (as inferred from Figure 6 of \cite{miltiadous2023dice}).

\begin{figure}[ht]
    \centering
    \begin{minipage}{0.5\textwidth}
        \centering
        \includegraphics[width=\textwidth]{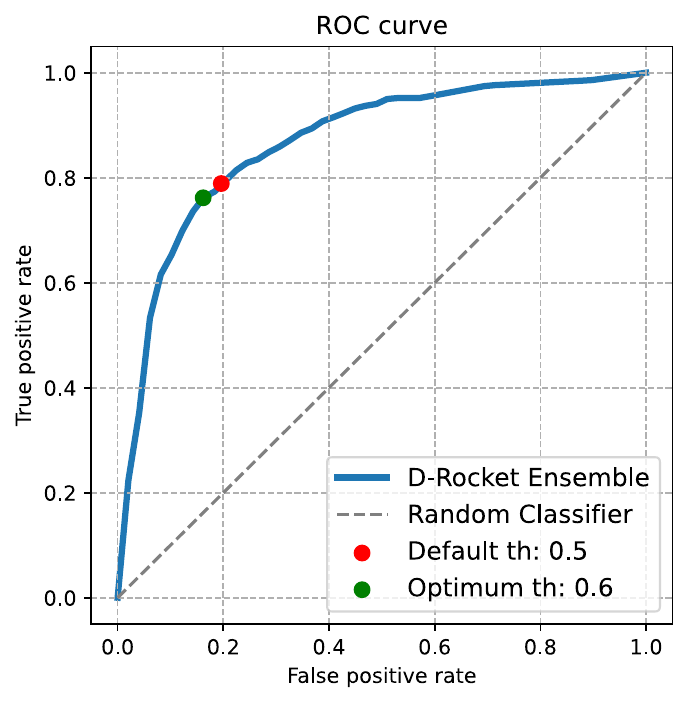}
        \centering
    \end{minipage}
    \hfill
    \begin{minipage}{0.4\textwidth}
        \centering
        \includegraphics[width=\textwidth]{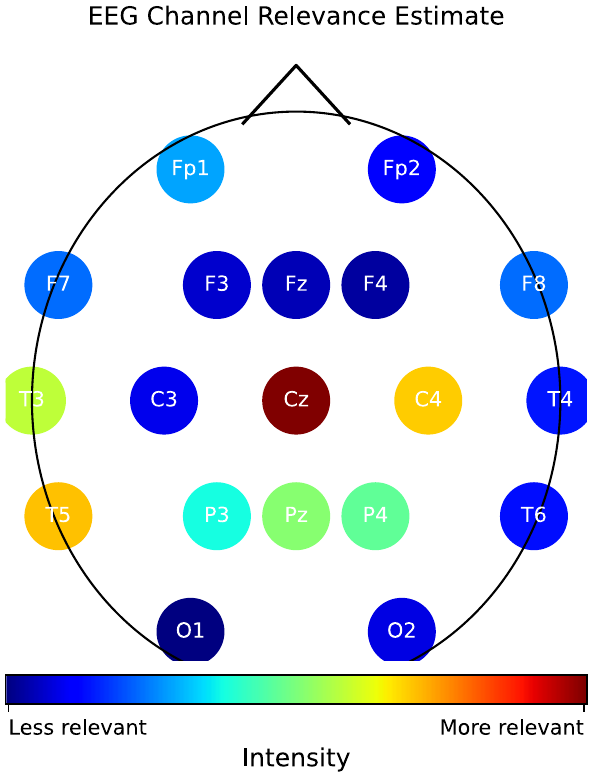}
        \centering
    \end{minipage}
    \caption{(Left) Receiver Operating Characteristic (ROC) curve obtained by applying different thresholds (th) over the class probabilities of each trial. (Right) Channel relevance of the Alzheimer's Disease classification task estimated by the Detach-Rocket ensemble.}
    \label{fig:AD_ROC_and_relevance} 
\end{figure}

In addition to these results, we also present the Receiver Operating Characteristic (ROC) curve in Figure \ref{fig:AD_ROC_and_relevance} (left), obtained by sweeping different thresholds over the predicted class probabilities. We highlight two points on the curve. The first one corresponds to the scores initially obtained with the default threshold of $0.5$. The second one shows the results obtained using the threshold that yielded the best accuracy, shown -along with the rest of the metrics- in Table \ref{tab:ad_cn_results}, which is slightly better than the one obtained with the default threshold ($79.86\%$ instead of $79.80\%$). Note that this analysis is possible thanks to the existence of a probability value for the class labels.

Finally, with the LOSO procedure, we obtain 65 different estimations of channel importance for this classification task. In Figure \ref{fig:AD_ROC_and_relevance} (right), we present the estimated channel importance averaged over the 65 folds. This figure demonstrates the potential of our methodology: without any prior feature engineering, we can identify the relevant brain areas for Alzheimer's disease classification in this dataset in a fully data-driven approach.



\section{Discussion}
Neuroscience applications typically involve multivariate time series data from a limited number of subjects with significant intersubject variability. This poses a challenge for deep learning classifiers, which can easily overfit the raw data and fail to generalize to subjects not included in the training set, creating a need for some form of feature engineering. In this context, simple ROCKET-based models have the potential to be an effective alternative, but they face some challenges regarding scalability with the number of channels and interpretability. This study aims to address these challenges by proposing Detach-Rocket Ensemble.

Similar to Arsenal, our proposed model is an ensemble of ROCKET-based models. As an ensemble, it has the advantages of reducing overfitting, being trivially parallelizable, and providing an intuitive label probability. However, the Detach-MiniRockets used in Detach-Rocket Ensemble are stronger based models, since they have a larger number of initial kernels and can better handle high-dimensional MTSC. In addition, after pruning, Detach-Rocket Ensemble ends up being a smaller model. For example, in the face detection challenge, the model was left with less than a third of the total kernels used by Arsenal.

One downside of the Detach-Rocket Ensemble is that the iterative detachment process must be conducted as many times as there are base models in the ensemble, creating an overhead in training time. However, this is compensated by the substantial reduction in the number of kernels during pruning, which makes the model require fewer convolutions during the forward pass, thus reducing inference time.

In future work, we plan to explore the iterative application of the Detach-Rocket Ensemble for channel pruning. After training the initial ensemble model, it is possible to use the estimated channel relevance to discard non-informative channels and then train a new ensemble model on the selected channels. This approach could improve model performance by providing a better coverage on the relevant channels. Additionally, it would be valuable to evaluate the Detach-Rocket Ensemble, using both Detach-MiniRocket and Detach-MultiRocket, on the UEA dataset to compare its performance with other non-ROCKET-based approaches.

\section{Conclusion}




In this study, we introduce Detach-Rocket Ensemble, an Multivariate Time Series Classification (MTSC) model that exploits the fast architecture of MiniRocket and the model size reduction provided by Sequential Feature Detachment (SFD) pruning. We demonstrate that Detach-Rocket Ensemble is able to handle both raw EEG and raw MEG data, achieving state-of-the-art performance while improving interpretability by providing built-in channel relevance and label probability.

Presented alongside a public repository with a user-friendly interface (\url{https://github.com/gon-uri/detach_rocket}), Detach-Rocket Ensemble represents a valuable resource for scientists working in the field of multivariate time series classification, particularly for neuroscience applications.


\begin{credits}
\subsubsection{\ackname} The authors would like to thank KTH Digital Futures and SciLifeLabs for granting their support in this project, as well as Federico Barone for engaging in insightful discussions.

\subsubsection{\discintname}
The authors have no competing interests to declare that are
relevant to the content of this article.
\end{credits}

%
%
%
\bibliographystyle{splncs04}
\bibliography{references}

\begin{thebibliography}{10}
\providecommand{\url}[1]{\texttt{#1}}
\providecommand{\urlprefix}{URL }
\providecommand{\doi}[1]{https://doi.org/#1}

\bibitem{alzubaidi2021review}
Alzubaidi, L., Zhang, J., Humaidi, A.J., Al-Dujaili, A., Duan, Y., Al-Shamma, O., Santamar{\'\i}a, J., Fadhel, M.A., Al-Amidie, M., Farhan, L.: Review of deep learning: concepts, cnn architectures, challenges, applications, future directions. Journal of big Data  \textbf{8},  1--74 (2021)

\bibitem{bagnall2018uea}
Bagnall, A., Dau, H.A., Lines, J., Flynn, M., Large, J., Bostrom, A., Southam, P., Keogh, E.: The uea multivariate time series classification archive, 2018. arXiv preprint arXiv:1811.00075  (2018)

\bibitem{bagnall2017great}
Bagnall, A., Lines, J., Bostrom, A., Large, J., Keogh, E.: The great time series classification bake off: a review and experimental evaluation of recent algorithmic advances. Data mining and knowledge discovery  \textbf{31},  606--660 (2017)

\bibitem{britton2016electroencephalography}
Britton, J.W., Frey, L.C., Hopp, J.L., Korb, P., Koubeissi, M.Z., Lievens, W.E., Pestana-Knight, E.M., St~Louis, E.K.: Electroencephalography (eeg): An introductory text and atlas of normal and abnormal findings in adults, children, and infants  (2016)

\bibitem{dau2019ucr}
Dau, H.A., Bagnall, A., Kamgar, K., Yeh, C.C.M., Zhu, Y., Gharghabi, S., Ratanamahatana, C.A., Keogh, E.: The ucr time series archive. IEEE/CAA Journal of Automatica Sinica  \textbf{6}(6),  1293--1305 (2019)

\bibitem{dempster2020rocket}
Dempster, A., Petitjean, F., Webb, G.I.: Rocket: exceptionally fast and accurate time series classification using random convolutional kernels. Data Mining and Knowledge Discovery  \textbf{34}(5),  1454--1495 (2020)

\bibitem{dempster2021minirocket}
Dempster, A., Schmidt, D.F., Webb, G.I.: Minirocket: A very fast (almost) deterministic transform for time series classification. In: Proceedings of the 27th ACM SIGKDD conference on knowledge discovery \& data mining. pp. 248--257 (2021)

\bibitem{dempster2023hydra}
Dempster, A., Schmidt, D.F., Webb, G.I.: Hydra: Competing convolutional kernels for fast and accurate time series classification. Data Mining and Knowledge Discovery  \textbf{37}(5),  1779--1805 (2023)

\bibitem{dietterich2000ensemble}
Dietterich, T.G.: Ensemble methods in machine learning. In: International workshop on multiple classifier systems. pp. 1--15. Springer (2000)

\bibitem{decoding-the-human-brain}
Emanuele, Mosi, P.A.: Decmeg2014 - decoding the human brain (2014), \url{https://kaggle.com/competitions/decoding-the-human-brain}

\bibitem{glover2011overview}
Glover, G.H.: Overview of functional magnetic resonance imaging. Neurosurgery Clinics  \textbf{22}(2),  133--139 (2011)

\bibitem{hamalainen1993magnetoencephalography}
H{\"a}m{\"a}l{\"a}inen, M., Hari, R., Ilmoniemi, R.J., Knuutila, J., Lounasmaa, O.V.: Magnetoencephalography—theory, instrumentation, and applications to noninvasive studies of the working human brain. Reviews of modern Physics  \textbf{65}(2), ~413 (1993)

\bibitem{henson2011parametric}
Henson, R.N., Wakeman, D.G., Litvak, V., Friston, K.J.: A parametric empirical bayesian framework for the eeg/meg inverse problem: generative models for multi-subject and multi-modal integration. Frontiers in human neuroscience  \textbf{5}, ~76 (2011)

\bibitem{ismail2020inceptiontime}
Ismail~Fawaz, H., Lucas, B., Forestier, G., Pelletier, C., Schmidt, D.F., Weber, J., Webb, G.I., Idoumghar, L., Muller, P.A., Petitjean, F.: Inceptiontime: Finding alexnet for time series classification. Data Mining and Knowledge Discovery  \textbf{34}(6),  1936--1962 (2020)

\bibitem{lawhern2018eegnet}
Lawhern, V.J., Solon, A.J., Waytowich, N.R., Gordon, S.M., Hung, C.P., Lance, B.J.: Eegnet: a compact convolutional neural network for eeg-based brain--computer interfaces. Journal of neural engineering  \textbf{15}(5),  056013 (2018)

\bibitem{middlehurst2021hive}
Middlehurst, M., Large, J., Flynn, M., Lines, J., Bostrom, A., Bagnall, A.: Hive-cote 2.0: a new meta ensemble for time series classification. Machine Learning  \textbf{110}(11-12),  3211--3243 (2021)

\bibitem{middlehurst2023bake}
Middlehurst, M., Sch{\"a}fer, P., Bagnall, A.: Bake off redux: a review and experimental evaluation of recent time series classification algorithms. arXiv preprint arXiv:2304.13029  (2023)

\bibitem{miltiadous2023dice}
Miltiadous, A., Gionanidis, E., Tzimourta, K.D., Giannakeas, N., Tzallas, A.T.: Dice-net: a novel convolution-transformer architecture for alzheimer detection in eeg signals. IEEE Access  (2023)

\bibitem{data8060095}
Miltiadous, A., Tzimourta, K.D., Afrantou, T., Ioannidis, P., Grigoriadis, N., Tsalikakis, D.G., Angelidis, P., Tsipouras, M.G., Glavas, E., Giannakeas, N., Tzallas, A.T.: A dataset of scalp eeg recordings of alzheimer’s disease, frontotemporal dementia and healthy subjects from routine eeg. Data  \textbf{8}(6) (2023). \doi{10.3390/data8060095}, \url{https://www.mdpi.com/2306-5729/8/6/95}

\bibitem{mizrahi2024comparative}
Mizrahi, D., Laufer, I., Zuckerman, I.: Comparative analysis of rocket-driven and classic eeg features in predicting attachment styles. BMC psychology  \textbf{12}(1), ~87 (2024)

\bibitem{ruiz2021great}
Ruiz, A.P., Flynn, M., Large, J., Middlehurst, M., Bagnall, A.: The great multivariate time series classification bake off: a review and experimental evaluation of recent algorithmic advances. Data Mining and Knowledge Discovery  \textbf{35}(2),  401--449 (2021)

\bibitem{rushbrooke2023time}
Rushbrooke, A., Tsigarides, J., Sami, S., Bagnall, A.: Time series classification of electroencephalography data. In: International Work-Conference on Artificial Neural Networks. pp. 601--613. Springer (2023)

\bibitem{schirrmeister2017deep}
Schirrmeister, R.T., Springenberg, J.T., Fiederer, L.D.J., Glasstetter, M., Eggensperger, K., Tangermann, M., Hutter, F., Burgard, W., Ball, T.: Deep learning with convolutional neural networks for eeg decoding and visualization. Human brain mapping  \textbf{38}(11),  5391--5420 (2017)

\bibitem{shifaz2020ts}
Shifaz, A., Pelletier, C., Petitjean, F., Webb, G.I.: Ts-chief: a scalable and accurate forest algorithm for time series classification. Data Mining and Knowledge Discovery  \textbf{34}(3),  742--775 (2020)

\bibitem{tan2022multirocket}
Tan, C.W., Dempster, A., Bergmeir, C., Webb, G.I.: Multirocket: multiple pooling operators and transformations for fast and effective time series classification. Data Mining and Knowledge Discovery  \textbf{36}(5),  1623--1646 (2022)

\bibitem{uribarri2023detach}
Uribarri, G., Barone, F., Ansuini, A., Frans{\'e}n, E.: Detach-rocket: Sequential feature selection for time series classification with random convolutional kernels. arXiv preprint arXiv:2309.14518  (2023), \url{http://arxiv.org/abs/2309.14518}

\bibitem{uribarri2023deep}
Uribarri, G., von Huth, S.E., Waldthaler, J., Svenningsson, P., Frans{\'e}n, E.: Deep learning for time series classification of parkinson's disease eye tracking data. arXiv preprint arXiv:2311.16381  (2023)

\bibitem{waytowich2018compact}
Waytowich, N., Lawhern, V.J., Garcia, J.O., Cummings, J., Faller, J., Sajda, P., Vettel, J.M.: Compact convolutional neural networks for classification of asynchronous steady-state visual evoked potentials. Journal of neural engineering  \textbf{15}(6),  066031 (2018)

\end{thebibliography}

\newpage
\section{Appendix}


\subsection{Face Detection hyperparameter optimization tables}

\vspace{-0.75cm} 
\renewcommand{\arraystretch}{1.2} 


\begin{table}[!ht]
\centering
\caption{Accuracies of several ROCKET variants on the validation set for different number of kernels. Both Detach-Rocket models use MiniRocket as the base model.}
\label{tab:kernel_optimization}
\begin{tabular}{|>{\raggedright\arraybackslash}m{5cm}|>{\centering\arraybackslash}m{1.6cm}|>{\centering\arraybackslash}m{1.6cm}|>{\centering\arraybackslash}m{1.6cm}|>{\centering\arraybackslash}m{1.6cm}|}
    \hline
    \textbf{Model} & \multicolumn{4}{c|}{\textbf{Accuracy (\%) for number of kernels}} \\
    \cline{2-5}
    & \textbf{1000} & \textbf{5000} & \textbf{10000} & \textbf{20000} \\
    \hline
    MiniRocket & 57.45 & 58.30 & 61.83 & 62.05 \\
    Detach-Rocket (single model) & 57.84 & 60.58 & 61.60 & 61.14 \\
    Detach-Rocket 10 model ensemble & 59.10 & 61.83 & \textbf{62.11} & 61.66 \\
    \hline
\end{tabular}
\end{table}

\vspace{-0.95cm} 

\end{document}